\documentclass[letterpaper, 10 pt, conference]{ieeeconf}  

\IEEEoverridecommandlockouts                              

\usepackage{latexsym}
\usepackage{amsmath}
\usepackage{booktabs}
\usepackage{graphicx}
\usepackage{color}

\usepackage{pifont}
\usepackage{algorithm}
\usepackage{algorithmic}
\usepackage{subcaption}
\usepackage{comment}
\usepackage{tikz}
\usepackage{graphicx}  
\usepackage{array}     
\usepackage{booktabs}  
\usepackage{url}
\usepackage{hyperref}

\newcommand{\themis}{\textsc{Themis}}

\newcommand*\emptycirc[1][0.8ex]{\tikz\draw[thick,draw=black] (0,0) circle (#1);}
\newcommand*\halfcirc[1][0.8ex]{%
  \begin{tikzpicture}
  \draw[fill=black!60!black,draw=black!60!black] (0,0)-- (90:#1) arc (90:270:#1) -- cycle ;
  \draw[thick,draw=black!60!black] (0,0) circle (#1);
  \end{tikzpicture}}
\newcommand*\fullcirc[1][0.8ex]{\tikz\draw[thick,draw=black!60!black,fill=black!60!black] (0,0) circle (#1);}

\title{\LARGE \bf
Themis: An explainable AI-enabled framework for Reinforcement Learning with Human Feedback
}

\author{Andreas Chouliaras$^{1}$ Luke Connolly$^{2}$ and Dimitris Chatzopoulos$^{1}$
\thanks{*This work is supported by EU Horizon project 101160671 (DIGITISE).}
\thanks{$^{1}$Andreas Chouliaras and Dimitris Chatzopoulos are with the School of Computer Science, University College Dublin, Belfield, Dublin 4, Ireland {\tt\small andreas.chouliaras@ucdconnect.ie}, {\tt\small dimitris.chatzopoulos@ucd.ie}}%
\thanks{$^{2}$Luke Connolly is an alumnus of the School of Computer Science, University College Dublin, Belfield, Dublin 4, Ireland {\tt\small luke.connolly@ucdconnect.ie}}%
\thanks{Published in © IEEE CAI 2026. Personal use of this material is permitted. Permission from IEEE must be obtained for all other uses, in any current or future media, including reprinting/republishing this material for advertising or promotional purposes, creating new collective works, for resale or redistribution to servers or lists, or reuse of any copyrighted component of this work in other works. DOI:\href{https://ieeexplore.ieee.org/abstract/document/11536497}{10.1109/CAI68641.2026.11536497}
}}

\begin{document}

\maketitle
\thispagestyle{empty}
\pagestyle{empty}

\begin{abstract}

Training safe Reinforcement Learning (RL) systems is inherently challenging, with no guarantee of avoiding unwanted behaviors. The most effective defenses against this are \textit{(i)} transparency through explainability and \textit{(ii)} alignment via human feedback. While both show promising results, no publicly available framework currently combines them. To address this, we introduce $\themis$, an XAI-enabled testing and evaluation framework for Reinforcement Learning from Human Feedback. $\themis$ supports over 200 widely used environments and is easily configurable for experiments in RL, transparency, and alignment. Our results show that $\themis$ can train reward models that match or outperform the environment’s true reward signal using human preferences. We also provide a cloud-based platform for collecting human feedback and managing experiments. It is user-friendly, auto-scalable, and supports large participant groups across multiple experiments without extra development overhead. Tests show $\themis$ can support one thousand users in back-to-back experiments on a modest commercial machine.

\end{abstract}

\section{INTRODUCTION}

Reinforcement Learning (RL) is a cornerstone of artificial intelligence~\cite{sivamayil2023systematic}, powering advances from autonomous vehicles~\cite{kiran2021deep} and robotics~\cite{tang2024deep} to industrial control~\cite{nian2020review} and large language models~\cite{ouyang2022rlhf}. Unlike supervised learning, RL agents learn through environment interaction and reward optimization. Deep RL (DRL) extends this by solving more complex tasks, though with higher computational and engineering costs~\cite{silver2016mastering}. This paradigm, however, raises a critical challenge: designing reward functions that \textit{accurately} capture intended behaviors. Poorly specified rewards cause \textit{reward hacking}, where agents exploit shortcuts to maximize returns but display unwanted behaviors~\cite{skalse2022defining, ibarz2018reward}. For instance, an RL taxi might speed to boost profits at the expense of safety. RL’s misalignments makes it fragile in safety-critical contexts.

To mitigate these issues, interactive Reinforcement Learning (iRL) leverages human knowledge, to improve agents before or during training. A prominent subset, Reinforcement Learning from Human Feedback (RLHF), enables agents to align with human values, needs and desires by iteratively refining their reward functions using feedback~\cite{christiano2017deep}. RLHF addresses three key challenges: \textit{(i)} resolving reward hacking by continuously correcting reward mispecifications~\cite{ibarz2018reward}, \textit{(ii)} accelerating training in problems with sparse rewards~\cite{wu2021human} and \textit{(iii)} enabling agents to learn complex behaviors that are difficult to define and program. Despite its newfound application to Large Language Models (LLMs), RLHF in other domains is limited by two systemic barriers.

\textbf{First}, existing RLHF frameworks lack generalizability and accessibility. Many works are proprietary, and open-source alternatives are often narrowly tailored to specific environments benchmarks(e.g robotics) limiting their utility for rigorous research. In both cases, these solutions are built on inflexible architectures that hinder customization. \textbf{Second}, the infrastructure for acquiring human feedback is fragmented. Existing platforms are dominated by the LLMs application case, offering little to no support for real-time interaction between participants and RL agents. Even the few that offer some support either lack scalability or provide inadequate tools for experiments with real humans.

Compounding these challenges is the under-explored role of explainable AI (XAI) techniques in RLHF. While XAI has been shown to improve human-AI collaboration in supervised learning by enhancing the users' situational awareness and the overall agent performance~\cite{paleja2021utility}, its potential in RLHF remains largely untapped. Explanations could, for instance, help human teachers diagnose agent failures during feedback cycles or reduce cognitive load by clarifying agent decisions. Yet, no existing RLHF framework integrates XAI techniques to amplify the effectiveness of human feedback~\cite{casper2023open}.

\begin{figure}[t]
    \centering
    \includegraphics[width=\columnwidth]{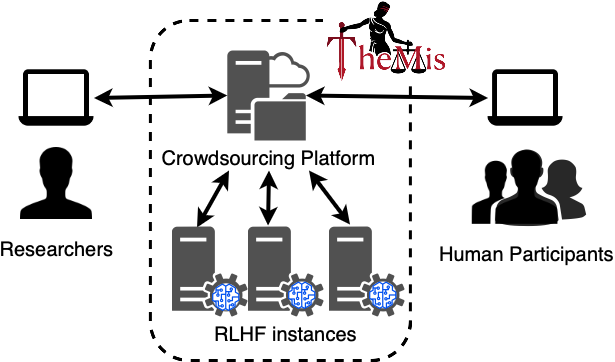}
    \caption{The $\themis$ framework and its interaction with researchers, human participants, and RLHF instances.\vspace{-0.5cm}}
    \label{fig:interaction_entities}
\end{figure}

In this work, we propose $\themis$\footnote{In ancient Greek mythology, $\themis$ was the goddess of divine order and judgment and associated with the scales of justice.}, a non-LLM focused RLHF framework\footnote{GitHub code: \url{https://github.com/achouliaras/Themis}} 
that enables experiments with real human participants or synthetic teachers and integrates: \textit{(i)} Reward learning and modeling, \textit{(ii)} incorporating XAI techniques with the RL agent, and (iii) a scalable cloud-based infrastructure for conducting and managing experiments~(Figure~\ref{fig:interaction_entities}). $\themis$ bridges critical gaps in RLHF research by:

\begin{enumerate}
    \item Supporting popular RL benchmarks (Atari, MuJoCo, Minigrid, BabyAI), and many customization options.
    \item Automatic algorithm selection (SAC, SAC-Discrete) and model configuration using CNN/MLP for discrete/continuous actions and tabular/pixel states.
    \item Reward modeling for RLHF experiments using environment rewards, real or synthetic human teachers.
    \item Ready-to-use XAI methods and support for custom.
    \item A cloud-based crowdsourcing platform for managing experiments via a unified web interface.
\end{enumerate}

$\themis$ includes an RLHF system and a crowdsourcing platform, connecting researchers, RL systems and human participants in a scalable way. Researchers can train multiple RLHF models with human feedback by simply logging in to the online crowd sourcing platform. Similarly, they also log in to the platform from a different interface as administrators, enabling them to oversee and manage multiple experiments simultaneously. Figure~\ref{fig:interaction_entities} illustrates this interaction.

We validate $\themis$' ability to distill human feedback to reward models by comparing preferences-based trained agents from synthetic teachers against environment reward trained agents. This way we can validate that reward modeling using preferences can correctly approximate the ground-truth reward. We also test the scalability of our crowdsourcing platform to handle multiple experiments and up to one thousand simulated human participants.

\smallskip
\noindent\textbf{Contributions.} In summary, in this work we:
\textit{(i)} identify the shortcomings of the works in RLHF and the lack of works that combine RLHF with XRL. Next, we \textit{(ii)} introduce an RLHF system with various features that make it easy to use and adapt and we test it for its validity as a human-feedback distillation tool. After that, we \textit{(iii)} introduce a crowdsourcing platform for acquiring human feedback as a cloud-based web platform and conduct multiple experiments to test its scalability and efficiency. Last, we \textit{(iv)} provide an open-source and highly configurable framework, named $\themis$, for experiments in iRL, RLHF and XRL. $\themis$~is highly configurable and can be easily deployed on multiple~devices. 


\section{BACKGROUND}\label{sec:background}
The RLHF paradigm is a learning procedure that requires careful decisions for all of its components: \textit{(i)} the environments, \textit{(ii)} the learning capabilities of RL algorithms and \textit{(iii)} agent modifications to enable learning from human feedback. Below, we discuss every aspect of such design choices, common for most RL(HF) tasks.

\smallskip
\noindent\textbf{Environment.} One broad category of design choices involves understanding the environment in which an agent operates. The agent receives an observation of the environment state at each timestep and interacts with it through specific actions. How we represent observations and actions significantly impacts the learning process and which RL algorithms are applicable. In some environments, the observation may be images, value arrays, or simple variable lists~\cite{sutton2018reinforcement}. Likewise, how the action space is defined is critical. Some environments use discrete action spaces—actions are chosen from a fixed list—while others use continuous spaces, where actions are vectors of floats. These allow infinite possible values, making the action space effectively infinite. Some RL algorithms work only with discrete actions, others with continuous, and some can be adapted to both~\cite{zhu2021overview}.

\smallskip
\noindent\textbf{RL agent.} In Deep RL, the internals of an RL algorithm (e.g., policy or value function) use neural networks as function approximators, depending on the state representation. States with geometric data like images benefit from computer vision methods such as convolutional neural networks (CNNs), while tabular states can be handled with Multi-Layer Perceptrons (MLPs). Another key aspect is how policies are learned. On-policy agents learn from their current behavior, while off-policy agents learn from other policies e.g., previous experiences or other agents~\cite{sutton2018reinforcement}. Off-policy methods are generally more sample-efficient and perform better in complex environments. In Deep RL, off-policy learning is typically enabled with a replay buffer (or experience memory), a data structure that stores past state transitions, actions, and rewards~\cite{mnih2015human}. At each timestep, models are updated using a sample from this buffer. Past experiences help generalization but increase resource usage.

\smallskip
\noindent\textbf{Reward function.} Even after all the above design choices a good reward function is crucial for good performance. The reward is what the RL agent optimizes and is the deciding factor of whether the agent will perform well in a particular task. A good reward function on a well designed RL system can lead to performance that exceed human capabilities~\cite{silver2016mastering}, though the danger of potential reward hacking is always there~\cite{skalse2022defining}. Instead of hand crafting the reward function, we can approximate them using ML models. Reward models depend on the data used. In Inverse RL, human demonstrations are used to train a reward function~\cite{abbeel2004apprenticeship}, working as a form of supervision in RL, but when demonstrations are hard to acquire, scientists turned to weaker forms of supervision.

\smallskip
\noindent\textbf{RLHF.}
Human feedback on an agent’s trajectory is the most common interaction method in iRL. Examples of interaction delivery include: i) binary facial expressions (happy/sad) on the agent’s current strategy~\cite{arakawa2018dqn}, and ii) preference elicitation between two strategies using binary choices~\cite{christiano2017deep,lee2021pebble,park2022surf} or non-binary rankings~\cite{cao2020human}. 
Ibarz et. al. combined demonstrations with human preferences by adding human-made trajectories to those evaluated via input. Human demonstrations were consistently preferred over agent-generated ones. The aim was to reduce reliance on human input, but this only works when demonstrations are feasible~\cite{ibarz2018reward}. Human feedback methods show strong results in user satisfaction and intuitiveness. They’ve been widely integrated into the learning process~\cite{cruz2020interactive} and support diverse input types with minimal impact on algorithmic agency~\cite{zhang2019leveraging}. Preference input is especially popular, backed by findings in psychology and social science on contrastive reasoning—humans naturally make decisions through comparison~\cite{hilton1986knowledge}.

\smallskip
\noindent\textbf{Explainable RL.} 
Even with a well-performing reward function or model, we can’t be sure it’s foolproof against reward hacking—especially since such failures are hard to detect and may occur under specific conditions. While integrating DL models into RL agents has greatly improved performance, it has also made their behavior more opaque, worsening this issue. Understanding the decision-making process through explanations is key to identifying weak policies and catching errors before deployment~\cite{heuillet2022collective, greydanus2018visualizing}. In RLHF, explainability allows human teachers to give more informed feedback, leading to safer and more effective agents. However, RL tasks differ from the supervised ML settings most XAI techniques come from. The temporal nature of RL and absence of ground truth makes explanation harder—but also open new channels unavailable in supervised ML~\cite{qing2022survey,vouros2022explainable}.

\section{Design of $\themis$} \label{themis}
\begin{figure}[t]
    \centering
    \includegraphics[width=\columnwidth]{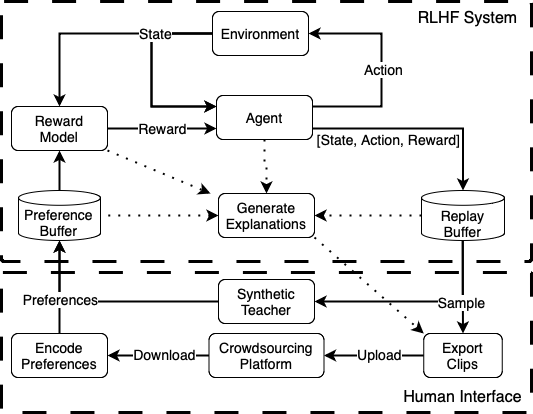}
    \caption{The $\themis$ framework portrayed as: i) the RLHF system that trains the reward model and the RL agent and ii) the Human Interface that provides the external API to connect with the crowdsourcing platform to acquire human feedback. The Generate Explanations module access any system parts needed by XRL methods.\vspace{-0.5cm}}
    \label{fig:framework}
\end{figure}

$\themis$ consists of two main subsystems (dashed boxes in Figure~\ref{fig:framework}): \textit{(i)} the RLHF system, which handles training, and \textit{(ii)} the human interface, which shows agent behavior examples to human participants and collects their preferences via the crowdsourcing platform. The RLHF system needs substantial computing power and memory, while the crowdsourcing platform requires high scalability. 

\subsection{RLHF System Design}

$\themis$ contains, apart from the classic agent-environment components, a reward model with a preference buffer, as depicted in Figure~\ref{fig:framework}. Testing can use either the crowdsourcing platform or synthetic teachers as in~\cite{lee2021bpref}.

\smallskip
\noindent\textbf{Parameters \& Experiment configuration.}
$\themis$ uses the Hydra framework~\cite{Yadan2019Hydra} to configure system parameters and all available learning hyperparameters for the RL agent, the reward model, explainable methods and the human interface. It saves the default configurations and modifications during execution in YAML files. This way multiple experiments can be conducted with minimal set up. 

\smallskip
\noindent\textbf{Environments.} $\themis$ uses the Gym API for its environments and supports most popular Gym environments like MuJoCo, Atari Games, Box2D~\cite{towers_gymnasium_2023}, Minigrid~\cite{MinigridMiniworld23} and BabyAI~\cite{chevalier2018babyai}. Each can be accessed through the environment domain and ID. It supports, pixel-based, tabular and grid-like state spaces, employing CNN networks along with MLPs when necessary. Both discrete and continuous action spaces are supported by automatic configuration of the RL agent in a simple and extensible way. Environment wrappers for further configurations are supported (e.g Gym, StableBaselines3) and custom ones can also be added.

\smallskip
\noindent\textbf{Agent \& Replay Buffer.}
The agent uses the Soft Actor-Critic (SAC) algorithm, a popular and well performing algorithm, modified to work on both continuous~\cite{haarnoja2017soft} and discrete action spaces~\cite{christodoulou2019soft}. SAC uses a replay buffer ($B$) for training by sampling transitions to train its value functions and to supply the reward model with data for retraining. The implementation follows StableBaselines3 by training two value functions simultaneously to reduce over-optimistic estimation of the state value. 
The agent also supports DL feature extraction for image state spaces and dense fully connected MLP Layers configured according to the action space the chosen environment uses.

\smallskip
\noindent\textbf{Reward Model \& Preference Buffer.}
Users can choose to use the environment reward or train a reward model from human preferences. Instead of the environment’s standard reward, we train a predictor model ($\hat{r}_\psi$) using an ensemble of three MLPs estimators $\psi$, trained on human-labeled instances to minimize prediction differences between the model and users. For testing, we also use synthetic teachers as in~\cite{lee2021bpref}. In more detail, given a segment $\sigma$, which is a sequence of state-action pairs $\{s_k,a_k,...s_{k+M},a_{k+M}\}$, the user exert their preference $y$ for two segments $\sigma^0$ and $\sigma^1$, with $y$ being the distribution of their preference ($[1,0]$ for the first, $[0,1]$ for the second and $[0.5,0.5]$ for equal preference). The results are saved in a preference buffer $D$ as $(\sigma^0, \sigma^1, y)$ and $\hat{r}_\psi$ is optimized using the Bradley-Terry model~\cite{bradley1952rank}: 

\begin{figure*}[t]
    \centering
    \includegraphics[width=0.95\textwidth]{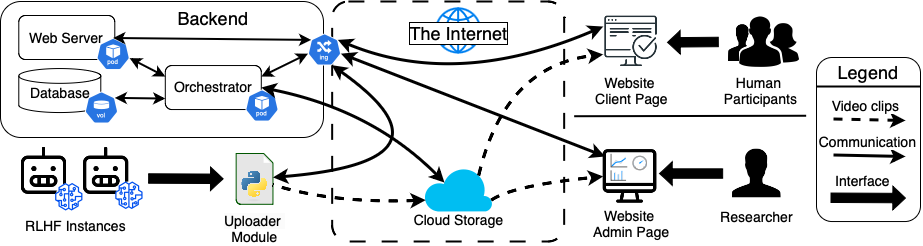}
    \caption{Crowdsourcing platform architecture. Using a website, researchers manage and oversee the experiments and the human participants can login and provide feedback. $\themis$' RLHF instances use the Uploader module to upload videos directly to a cloud storage and receive the human preferences. The smooth interaction and scaling of the platform is handled by the backend, hosting the web server and the database.}
    \label{fig:overview of architecture}
\end{figure*}

\begin{equation*}
    P_{\psi}[\sigma^1 \succ \sigma^0] = \frac{exp \sum \hat{r}_\psi(s^1_t,a^1_t)}{ exp \sum \hat{r}_\psi(s^0_t,a^0_t) + exp \sum \hat{r}_\psi(s^1_t,a^1_t) }
\end{equation*}

where $P_\psi[\sigma^i \succ \sigma^j]$ is the probability that $\sigma^i$ is more preferable than $\sigma^j$ using estimator $\psi$.
$\hat{r}_{\psi}$ is optimized to minimize the cross-entropy loss between the predictions and the actual human preference:
\begin{multline}
    J_{\hat{r}} = - \sum_{(\sigma^1,\sigma^2,y)\in D} y(0)log P_\psi[\sigma^0 \succ \sigma^1] \\
    + y(1)log P_\psi[\sigma^1 \succ \sigma^0]
\end{multline}
The purpose of the reward model is to reduce the interaction time required by the users. Christiano et. al. has previously demonstrated that using a preference predictor as a reward model can reduce human interaction by an order of 1000 without any noticeable difference in performance~\cite{christiano2017deep}. Furthermore, we provide a customizable linear reward scaling function defined as $f(\hat{r}) = \alpha * \hat{r} + \beta$, so that the rewards produced by the model has the same scale as the environment's rewards. The true reward is also saved for comparisons and plotting and can even be combined with the human generated reward if desired. In situations however, where the true reward is not provided or cannot be easily discerned the human input can be used as the sole reward for the agent to learn. The user can specify various parameters regarding the reward model training, some associated with the frequency of the interactions, the number of clips generated in each interaction phase but also the size of the clip segments, their freshness and even the rules for sampling them and organising them into pairs. The methods available to sample and generate segment pairs are Maximum Disagreement, Entropy, and K-Centered or custom. Sampling is important, especially when annotation budgets are constrained~\cite{chouliaras2025maximizing}.

\subsection{Human Interface} 
This library provides methods to convert paired segments to video clips, receive human input, and generate explanations. It is comprised of three modules: (a) Export Clips, (b) Encode Preference, and (c) Generate Explanations. 

\smallskip
\noindent\textbf{Export Clips}. This module provides methods that convert the state-action pairs to frames and export them as short videos. If an explanation method is used that creates visualisations, these methods can be used to overlay or concatenate them to the relevant video segments. As the state and action data saved are not usually in a form the user can understand, we provide methods that use them to render viewable frames from which the video clips can be generated. In more detail, we employ a simulation copy of the training environment with the ability to change the internal state of the environment to jump to each state of each segment and then proceed to render it into frames. If the environment is not deterministic, we need to change the environment's internals for every state in the segment. Otherwise, we can jump at the initial state of the segment and use the pre-selected actions in the segment to render the frames without forcibly changing the environment internals for each frame. Integration with the crowdsourcing platform uses the Uploader Module.

\smallskip
\noindent\textbf{Encode Preference.}
This module receives user input, encodes it into the reward model’s format, and stores it in the preference buffer. The Uploader Module is used to retrieve feedback from the crowdsourcing platform.

\smallskip
\noindent\textbf{Generate Explanations.} This module is associated with the feedback acquisition functionality of the reward model by human teachers and is responsible for generating explanations on the segments before the user is asked to label them. Based on which part of the RL system the intended explanation method applies, multiple system parts can be accessed to generate the explanations. It is also possible to evaluate the effect of combining explainable techniques as in~\cite{huber2021local}. We identify three potential areas of the RL systems that explanations methods could be used: 
(a) saliency map methods which can be applied in the actor or critic to extract local explanations for the policy or the value function, (b) policy summarization methods that can be used to extract global explanations by accessing the replay buffer and the actor \& critic models, and (c) methods that explain the agent's behavior through the reward it receives by accessing the reward model or the environment's true reward. 

In our implementation we provide three plug-n-play explainable methods using the PyTorch library Captum~\cite{kokhlikyan2020captum}: (i) \emph{Integrated Gradients}~\cite{sundararajan2017axiomatic}, (ii) \emph{Kernel SHAP}~\cite{lundberg2017unified}, (iii) \emph{TracInCP}~\cite{pruthi2020estimating}.

\subsection{Crowdsourcing platform design} 
Our architecture is divided into \emph{(a) the front-end}, and \emph{(b) the back-end}. Figure \ref{fig:overview of architecture} shows the platform architecture. 

\smallskip
\noindent\textbf{The Front-end.} Our crowdsourcing platform involves three entities: (a) the requesters that initiate the tasks, (b) the workers that complete these tasks and (c) the moderators that oversee the whole process. In this setting, the requesters are multiple instances of $\themis$ that uses the modules: Export clips, Encode Preferences and Uploder to export video clips, upload them to the crowdsourcing platform, receive the preferences from human participants and save them locally for training. The workers are human participants in such experiments that log into our platform and view clips and provide their preferences. The moderators are the researchers that log into our platform as admins and can determine when and which of the uploaded videos are released to the crowd and which ones from the completed samples are returned back to $\themis$ to continue training. Through the platform the researcher has the ability to adjust several experimental parameters associated with the crowdsourcing procedure like how many inputs a single participant can provide and how many times a pair of video clips will be displayed. The crowdsourcing platform is hosted by the \emph{Web Server} that runs in the \emph{Back-end}.

\smallskip
\noindent\textbf{The Uploader module.} The functionality necessary to upload video clips and download the feedback is automated through this Python module. These methods are called in the export and encode preference modules. For each video needed to be uploaded, the Uploader first requests from the Web Server an authorised URL to upload them to the Cloud-based blob storage. Then, the web server request the cloud storage to create a unique authorized link for a file upload. Then, web server returns the authorized link to the Uploader module and finally it uploads the video clip directly to the cloud storage using the authorized link. The whole interaction is secured through API keys that verify that the authorised upload URLs are handled only by the authorized entities. These interactions are depicted in Figure \ref{fig:overview of architecture}.

\smallskip
\noindent\textbf{The Back-end.} The entire platform is built on Docker containers and we use Kubernetes to efficiently manage the scalability needs of the platform. We use an orchestrator to facilitate communications between the instances of our \emph{Web Server} and our \emph{database}, while the use of Kubernetes Ingress provides a single communication interface with the outside network. This enables the platform to operate smoothly when the number of users or experiments increase and the containers for the underlying containers replicate to meet the increased load. 
Moreover, the \emph{database} stores the feedback acquired by the human participants for all experiments in operation, until they are given back to the RLHF systems in which case they are no longer needed, and are discarded. For our \emph{database} we use a NoSQL database through MongoDB~\cite{mongodb} to save the IDs of pairs of video clips and the preferences of the users. The choice for a NoSQL database was in purpose, in order to provide flexibility for other feedback methods or saving more data per video clip or per pair. It also supports saving metadata that is helpful for experiment management. To minimize the storage required by the crowdsourcing platform we do not save the actual video clips in our \emph{database} and instead we employ a cloud-based blob storage (Microsoft Azure in our case). Whenever the Uploader module requests to upload a video clip, we create an entry in our \emph{database} with the related data, we give the video a unique ID and request our blob storage to also make an entry there and return an authorized upload URL that is handed over to the Uploader module to upload the video.
Lastly, the \emph{Web Server} hosts the crowdsourcing platform and was developed using Next.js~\cite{nextjs}. It communicates with the \emph{database} and the cloud storage to load the videos that will be displayed to the users and then stores their preference to the database. 


\section{Experiments and Results} \label{sec:exp}

\noindent\textbf{Setup.} For our experiments, we deploy the RLHF system in a enterprise level computer server to validate the training process with the following characteristics: an Intel Xeon 6240 processor with 18 physical cores clocked at 2.1 GHz, 384 GB RAM and an Nvidia Tesla V100 graphics card with 32GB VRAM. To test the performance of our crowdsourcing platform we deploy it in a commercial computer system with the following characteristics: an AMD Ryzen 5 processor with 6 physical cores and 35 MB of cache memory, 16 GB DDR4 memory, an NVMe SSD drive with reading speed of 3500MB/s and writing speed of 3000MB/s, and an NVIDIA GeForce RTX 3060 Ti graphics card with 8GB of memory. The network connection at has $\approx 500$ Mbps download speed, $\approx 50$ Mbps and $\approx 3$ ms of latency. 

\begin{figure*}[t]
    \centering
    \includegraphics[width=\textwidth]{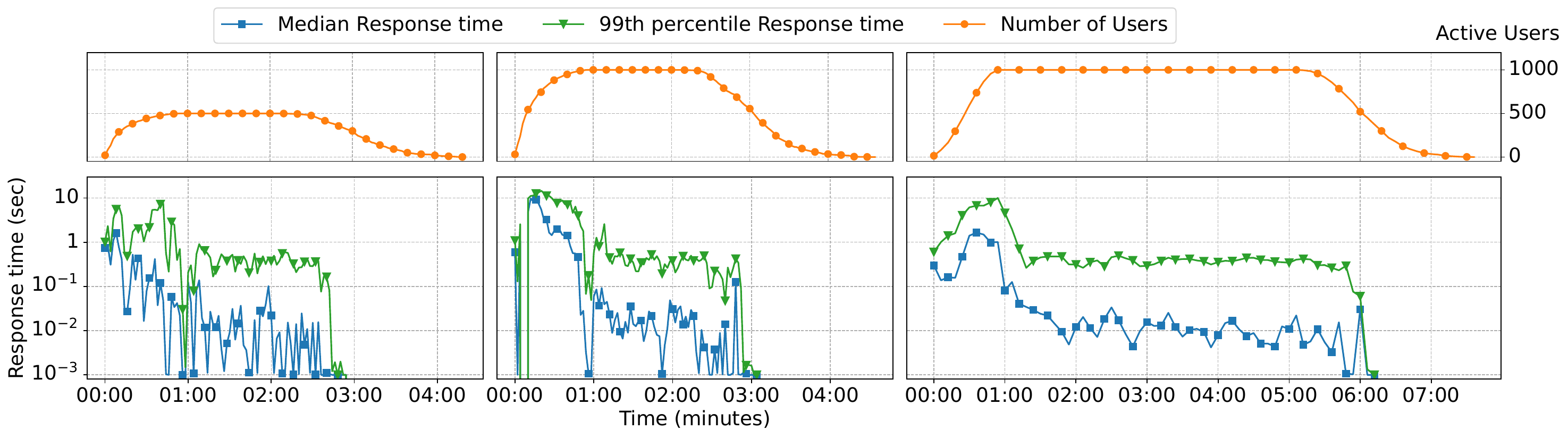}
    \caption{Median \& 99th percentile response times based on the number of active users. The total number of users are equally split into two groups with slightly different viewing patterns that are connecting from different locations.}
    \label{fig:exp_web_response}
\end{figure*}

\begin{figure}[t]
    \centering
    \includegraphics[width=1\columnwidth]{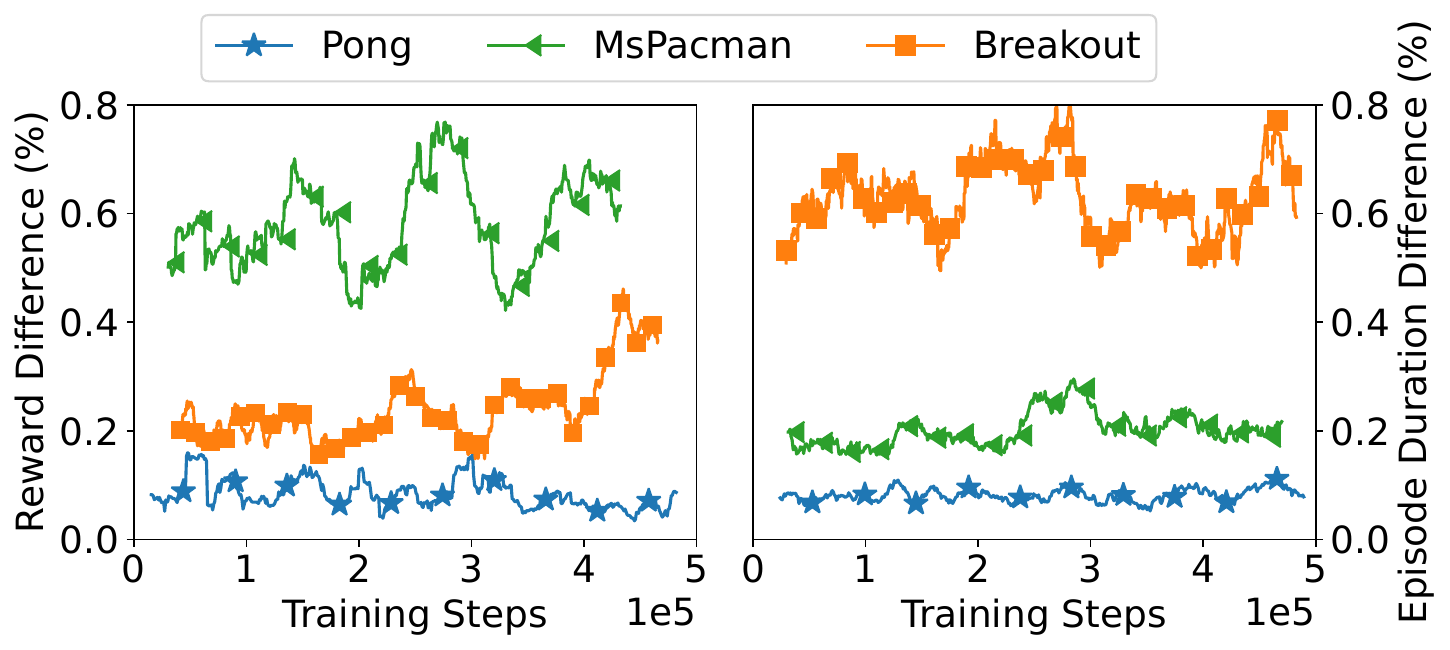}
    \caption{
    Agent's difference (\%) on reward when trained using true reward vs reward model (left) and episode duration (right) for various environments. The positive values indicate the reward model outperforms the reward function.
    }
    \label{fig:exp}
\end{figure}

\noindent\textbf{Rationale of RL experiments.} To validate that the RLHF system in $\themis$ is able to train good reward models using preferences, we train reward models using preferences from synthetic teachers and compare it against agents that use the environment's true reward signal. The synthetic teachers indicate preference based on the higher true cumulative reward in the segments. In this way, the reward models on the experiments will be evaluated on their learning ability to approximate the true reward function. We use the true environment reward as the metric to compare the agents trained on the reward function with the agents trained with a reward model. We conduct our experiment on three very popular environments from the Arcade Learning Environment~\cite{bellemare2013arcade}: Pong, MsPacman and Breakout. The results are depicted in Figure~\ref{fig:exp}. On the left figure, we display the percentage difference of the rewards from training the agent using the true reward versus using the reward model and in the right figure, we display the percentage difference in the episode duration when training the agent using the true reward vs using the reward model. Both figures use moving averages across multiple episodes to smooth out the variances and reveal trends. Positive values indicate that the agents trained using the reward model perform better than agents trained with the reward function from the environment. Based on Figure~\ref{fig:exp}, we can see that $\themis$  trains reward models that are good estimators of the true reward. In some cases, the reward model even manages to outperform the usage of the true reward function (upwards trend of reward in MsPacman and Breakout), indicating that reward models also benefit from reward shaping, providing regular rewards compared to reward functions that are sparser. We can also observe some big drops in reward differences for the MsPacman environment. These drops always seem to occur after the reward model is retrained. This phenomenon could be attributed to a potential domain shift in the reward model after its retaining and the agent needs to adapt to the updated reward model before it starts displaying even greater performance than before. This is evident in MsPacman as it requires the agent to develop strategic behaviours to not only survive for longer but also score more points in the meantime. In general, it seems that the agents trained using reward models are able to get on average higher true cumulative rewards and interact with the environments for slightly longer. We ought to mention that these results are invariable to the usage of our supported explanation techniques, only increasing the duration of the overall experiments. 

\begin{figure}[t]
    \centering
    \includegraphics[width=1\columnwidth]{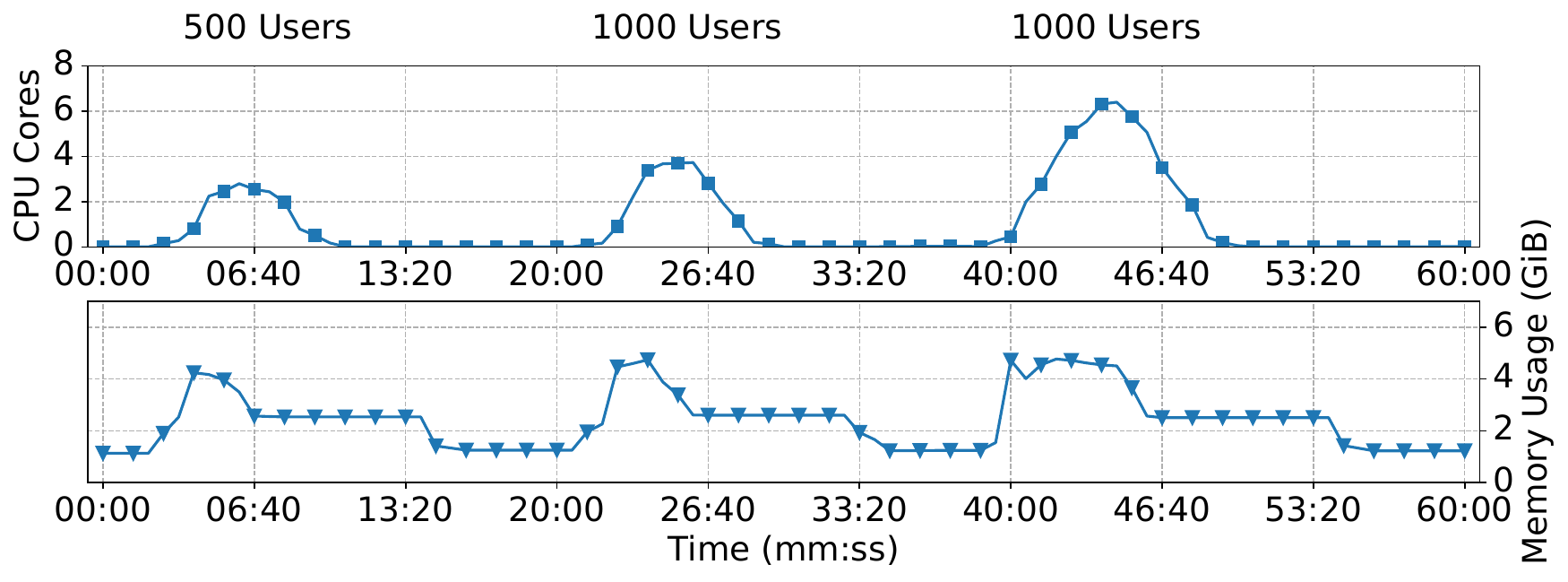}
    \caption{Crowdsourcing platform median response time on increasing number of active users to 200, 500, and 1000.}\vspace{-0.5cm}
    \label{fig:exp_pc_response}
\end{figure}

\smallskip
\noindent\textbf{Rationale of experiments for human feedback.} To test the capabilities of our platform we employ the Loadster app~\cite{loadster} to create bots that login to our web app, view videos and provide feedback, following a scripted behaviour. This simulates the load of the system at the experiment start, during peak load, and at the experiment finish. We report the 99th percentile and median response times of the scripted users in comparison to the currently active number of users. We conducted three experiments: i) 500 users with one-minute aggressive ramp up, one-minute peak load and two-minute smooth ramp down, ii) 1000 users with one-minute aggressive ramp up, one-minute peak load and two-minute smooth ramp down and iii) 1000 users with a one-minute smooth ramp up, four minutes peak load and two-minute smooth ramp down. We split the users into two equal groups with slight variations in their scripts regarding the viewing duration of the videos. This simulated conducting controlled experiments with multiple groups at the same time. Group 1 connected from Virginia, USA and Group 2 from Oregon, USA, while the Web server was deployed in our lab. As depicted in Figure \ref{fig:exp_web_response}, in all three experiments, we observe longer response times at the first minute of the experiment, which soon stabilizes as we reach the maximum number of active users. During peak load, the median response times fall below 0.1 seconds, dropping even further when the users start to log off. Extending the peak load duration of the experiment from one minute to four minutes does not have any considerable effect on the response time (Figure~\ref{fig:exp_web_response}) and the memory usage but leads to an increase in the number CPU core usage to almost double (Figure~\ref{fig:exp_pc_response}). 

In RLHF experiments, the number of active users rapidly increases in the beginning, as all participants try to login into the web platform. Gradually, it reaches to the maximum number of active users for the duration of the experiment and then gradually falls as the numbers gradually log off the platform. As displayed in Figure \ref{fig:exp_web_response}, the greatest strain in our system is at the start of the experiments as the system components replicate and scale up and accommodate to the need for additional resources due to the increasing number of users. As each user needs different amount of time for viewing videos and make decisions, the computational load and memory requirements stabilize (Figure \ref{fig:exp_pc_response}) and the response times drops to below one second (Figure \ref{fig:exp_web_response}). During the experiment under max user load, median response time never exceed 0.1 seconds, ensuring smooth experiments.

Notably, the cost of deploying our crowdsourcing platform to a popular cloud service is very low. In detail, considering 1000 users and an experiment duration of less than an hour, the following options are sufficient: a) a c6a.xlarge instance from Amazon Web Services (4 vCPUs, 8 GiB RAM) at \$0.153 per hour, b) a c2-standard-4 instance from Google Cloud Platform (4 vCPUs, 16 GiB RAM) at \$0.1816 per hour and c) for a F4s v2 instance from Microsoft Azure (4 vCPUs, 8 GiB RAM) at \$0.169 per hour. The cost and duration of training agents with the RLHF system however, is highly dependent on the accelerator used. 


\section{Related Work}\label{sec:relatedwork}
In this section we discuss the state-of-the-art for (RLHF) and review works that are methodologically similar to ours.
The first major work in the area of RLHF was from Christiano et. al.~\cite{christiano2017deep} that used the MuJoCo environments~\cite{todorov2012mujoco} and the Arcade Learning environment~\cite{bellemare2013arcade}. They experimented with both continuous and discrete action spaces but using different RL algorithms for each case that are not in the state-of-the-art (Trust Region Policy Optimization (TRPO) algorithm for continuous environments~\cite{schulman2015trust} and the Advantage Actor Critic (A2C) algorithm for discrete environments~\cite{mnih2016asynchronousmethodsdeepreinforcement}). Furthermore, although accessible, their implementation has the reward model and the human feedback acquisition platform separate from the RL agent training framework and its environments, not being able to function as a complete testing and evaluation framework, especially in the case of human-in-the-loop systems that train reward models more that one time.  Later Ibarz et. al. presented a work that relies on both human demonstrations and human feedback to train RL agents~\cite{ibarz2018reward}. This work stands as a direct successor of \cite{christiano2017deep} and although it makes interesting findings, their implementation is not publicly available, preventing usability by other researchers.


\begin{table}[t]
    \footnotesize
    \begin{tabular}{%
    p{12.0em}
    >{\raggedright\arraybackslash}p{1.5em}
    >{\raggedright\arraybackslash}p{1.2em}
    >{\raggedright\arraybackslash}p{1.2em}
    >{\raggedright\arraybackslash}p{0.2em}
    >{\raggedright\arraybackslash}p{1em}
    >{\raggedright\arraybackslash}p{1em}}
    & \rotatebox[origin=b]{25}{\textbf{Christiano et. al.}}
    & \rotatebox[origin=b]{25}{\textbf{Ibarz et. al.}}
    & \rotatebox[origin=b]{25}{\textbf{PEBBLE}}
    & \rotatebox[origin=b]{25}{\textbf{BPref}}
    & \rotatebox[origin=b]{25}{\textbf{EXPAND}}
    & \rotatebox[origin=b]{25}{\textbf{Themis}} \\ 
    \toprule 
        Continuous Action Spaces & \fullcirc & \emptycirc & \fullcirc & \fullcirc & \emptycirc & \fullcirc \\    
        Discrete Action Spaces & \fullcirc & \fullcirc & \emptycirc & \emptycirc & \fullcirc & \fullcirc \\
        Modern RL Algorithm & \emptycirc & \emptycirc & \fullcirc & \fullcirc & \emptycirc & \fullcirc \\
        \midrule
        Gym MuJoCo Support& \fullcirc & \emptycirc & \fullcirc & \fullcirc &  \emptycirc & \fullcirc \\
        Gym Atari (ALE) Support& \fullcirc & \fullcirc & \emptycirc & \emptycirc & \halfcirc & \fullcirc \\
        Minigrid \& BabyAI Support& \emptycirc & \fullcirc & \emptycirc & \emptycirc & \halfcirc & \fullcirc \\
        \midrule
        Supports Synthetic Teachers & \emptycirc & \fullcirc & \fullcirc & \fullcirc & \emptycirc & \fullcirc \\ 
        Supports Human Teachers & \fullcirc & \emptycirc & \fullcirc & \emptycirc & \fullcirc & \fullcirc \\ 
        Open Source Code & \halfcirc & \emptycirc & \halfcirc &  \fullcirc & \emptycirc & \fullcirc \\
        Crowdsourcing Platform & \fullcirc & \emptycirc & \emptycirc & \emptycirc & \emptycirc & \fullcirc \\
        XAI Plug-in Enabled & \emptycirc & \emptycirc & \emptycirc &  \emptycirc & \emptycirc & \fullcirc \\
        \bottomrule
        \end{tabular}
        \caption{Comparison of our framework with related works. \fullcirc/\emptycirc/\halfcirc:~ Presence / Absence / Partial presence of the feature. \vspace{-0.5cm}}        
        \label{tab:comparison}
\end{table}
Lee et al., on the other hand, in their two works (PEBBLE~\cite{lee2021pebble} and BPref~\cite{lee2021bpref}) employ Soft Actor-Critic (SAC)~\cite{haarnoja2017soft}, which is a more modern algorithm compared to TRPO and A2C Christiano et. al. use. Still, they only support robotic tasks with continuous action spaces like MuJoCo~\cite{todorov2012mujoco}. Moreover, BPref only trains reward models with synthetic teachers, and it does not provide the necessary infrastructure to conduct experiments with real humans. 

Then, Guan et. al. use with EXPAND~\cite{guan2021widening} a different approach for the user to provide feedback for training the RL agent in the form of binary feedback (Good, Bad) and through interactive visual explanations. They do not train a reward model, having the user directly interact with the learning algorithm. This method is very labor intensive as human teachers need enormous numbers of interactions to train an RL agent successfully. The participants need to provide feedback that contains a lot of information and requires careful thinking, making it more prone to misalignment or disagreement in the feedback they provide than preference-based methods. Although this work is also trying to bridge XRL with RLHF, their work is limited to discrete action spaces, is not open-source and is not focused on providing explanation to humans but the exact opposite.

Compared to $\themis$, the works above lack necessary tools for conducting RLHF experiments across different domains and offer no explainability support (Table~\ref{tab:comparison}).

\section{Conclusion \& Future Work} \label{sec:future}
In this work, we presented $\themis$, a highly configurable testing and evaluation framework for RLHF with explainability support. We also incorporate a web-based microservice application that scales in a cost-efficient way and can support experiments with multiple groups of users, tested with up to one thousand participants. We believe $\themis$ will be helpful to the research community and a valuable addition to the body of open-source tools. We invite the research community to use $\themis$ for their research.

In the future, we plan to integrate more semi-supervised and self-supervised RL methods to offer more options on efficiently jump-starting human feedback acquisition without the short and iterative human-in-the-loop processes that most works have done in the past. Understanding the effects of pretraining methods in RLHF in terms of performance, stability, scalability, and feedback quality has the potential to bring interesting advancements. We also believe that the reward model is a component worth explaining in greater detail, to help understand the mechanisms of human feedback distillation to rewards. Any insights made on these mechanisms can lead to improvements in human feedback quality. 


                                  
\newpage
\bibliographystyle{IEEEtran}
\bibliography{xirl,tools_repos}

\section*{APPENDIX A: Using $\themis$}
In this section, we explain how to set up $\themis$, deploy it, and use it to conduct experiments with synthetic teachers and human participants. We separate the procedure into five phases: I- deploying the crowdsourcing platform, II- deploying the pre-training of the RLHF system, III- acquiring human feedback, IV- deploying the training of the RLHF system and V- analyzing the results.

\smallskip\noindent\textbf{I --  Crowdsourcing platform deployment.}
To setup the crowdsourcing platform for the first time we need the following tools: i) an active CloudFlare Tunnel, ii) configure Clerk for user authentication, iii) set up Azure Blob Storage, iv) configure Resent for email notifications. After accounts are created and the relevant API keys are issued they can be filled to the configuration files of our platform. We then deploy our platform in K3s to the desired device (our Github repository contains a fully detailed setup guide).
To run an experiment we deploy the crowdsourcing platform to work as a server. Any RLHF instance can communicate with it to send videos and receive feedback via the Uploader module. 
RLHF instances can be deployed in the same device or separately; in either case they can all communicate with the crowdsourcing platform, supporting multiple parallel experiments.

\smallskip\noindent\textbf{II -- RLHF pre-training.} 
In this phase, the RLHF system trains the agent in an unsupervised manner, trying to maximise an intrinsic reward $r^{int}$ calculated from the states' entropy in a trajectory as in~\cite{lee2021pebble}. The generated episode trajectories are saved in the replay buffer $B$ (Appx. \ref{appendixA}, Algorithm~\ref{alg:pre}) and are used to jump-start the agent's and reward model's learning in the training phase. At the end of the pre-training phase, the RLHF system saves the agent and the reward model in files to serve as checkpoints before training phase.

\smallskip\noindent\textbf{III -- Human feedback acquisition.} The RLHF system deploys the agent in the simulation environment, samples trajectories, and couples segments of equal size based on the sampling rule chosen, e.g. maximum disagreement. It then, generates video clips that are uploaded to the crowdsourcing platform for feedback acquisition. If synthetic teachers are used, this last step is omitted, and instead, the preference is noted based on the sum of true rewards in each segment. In this case, the crowdsourcing platform is not engaged. Otherwise, the RLHF system waits for feedback and using our crowdsourcing platform the researcher configures the experimental parameters of all crowdsourcing tasks and enables access to the video clips from the participants. The participants login to the platform, watch the videos and provide feedback until all videos gain the specified amount of feedback. When the experiment is completed the researcher can inspect the results before the platform returns the feedback back to the RLHF system(s) and save them in their preference buffer.

\smallskip\noindent\textbf{IV -- RLHF training.}
In this phase, we first reset the critic networks to learn the new reward function using the human preferences. This is done by sampling labelled pairs from the preference buffer and training the reward model for a fixed number of iterations. With the reward model trained from the human feedback we deploy the RL agent to the environment using the reward model to provide rewards for its actions. The agent picks an action $a_t$ based on the current state $s_t$ under policy $\pi_\phi$, records the transition to $s_{t+1}$ and receives the reward $r_t$. This reward, however, is not the one that the environment may provide but the output from querying the reward model by providing it with the current state and action selected. Then, as in the pre-training phase, the transition is stored in the replay buffer $B$ and proceeds to the gradient update step of the agent. Because we are using SAC, this involves updating the critic, the actor, and the temperature term $\alpha$ and periodically, we also apply a soft update to the target critic network. The training algorithm is depicted in Appx. \ref{appendixA}, Algorithm~\ref{alg:train}. When the RL agent trains for the specified number of steps, it stops, saving the agent models and data structures. Then, it is deployed in the evaluation environment, measuring its performance.

\smallskip\noindent\textbf{V -- Data Analysis.}
Steps III and IV can be repeated multiple times, further refining the reward model and the agent. 
When the reward model and the agent converge, and the learning stops, the researcher can access the logs to analyse the experiment results. Multiple metrics can be used to investigate the overall performance of the agent (e.g. episode length, cumulative reward, actor and critic loss). The whole experimental scenario can be repeated and modified depending on the scope. This can result in multiple agents that are trained from different experimental conditions and can be directly compared from their performance on the environment. Explanation techniques can also be used to provide supplemental information to the human participants about the agent's reasoning, potentially leading to better feedback and better agent performance on the environment.

\section*{APPENDIX B: Algorithms}
\label{appendixA}

The training algorithm for Themis (Alg. \ref{alg:train}) combines the Soft Actor-Critic (SAC) RL algorithm from Haarnoja et. al in~\cite{haarnoja2017soft} with the RLHF approach introduced by Christiano et. al. in~\cite{christiano2017deep}. We also employed Christodoulou's adaptation of SAC for discrete action spaces~\cite{christodoulou2019soft}. As for, the pretraining algorithm (Alg. \ref{alg:pre}), we use intrinsic rewards based on the state entropy as in~\cite{lee2021pebble} to kickstart the RL agent training. Other pretraining approaches can also be added. 

\begin{algorithm}[ht]
    \caption{\themis, Interactive Training Phase} 
    \label{alg:train}
    \begin{algorithmic}[1]
        \STATE Initialise environment, learning parameters, buffer $D$, temperature $\alpha$ and networks $\theta_1,\theta_2, \hat{\theta}_1, \hat{\theta}_2$
        \STATE Load buffer $B$, network $\phi$, model $\hat{r}_\psi$ for $\psi \in [1,3]$
        \FOR{each episode}
            \STATE Reset environment and get $s_0$
            \FOR{each timestep $t$} 
                \STATE $a_t \sim \pi_\phi(a_t|s_t)$
                \STATE $s_{t+1} \leftarrow \mathcal{P}(s_t,a_t) $ \COMMENT{Sample transition from env}
                \STATE $r_t \sim \hat{r}_\psi(s_t, a_t)$
                \STATE $B \leftarrow B \cup \{(s_t,a_t,s_{t+1},r_t)\}$
                \IF{step \% $k$ == 0}
                \FOR{$ i= 1,2, \dots,M $} 
                    \STATE $(\sigma^0,\sigma^1) \sim B$
                    \IF{human\_teacher}
                        \STATE $(\chi_0,\chi_1) \leftarrow gener\_explanations(\sigma^0,\sigma^1)$
                        \STATE $\nu \leftarrow export\_clips(\sigma^0,\sigma^1,\chi_0,\chi_1)$
                        \STATE $y \leftarrow query\_human\_teacher(\nu)$
                    \ELSE
                        \STATE $y \leftarrow query\_synthetic\_teacher(\nu)$
                    \ENDIF
                    
                    \STATE $D \leftarrow D \cup \{(\sigma^0,\sigma^1,y)\}$
                \ENDFOR
                \FOR{each gradient step} 
                    \STATE sample minibatch $\{(\sigma^0,\sigma^1,y)_j\} \sim D$
                    \STATE Optimize $J_{\hat{r}_\psi}$ for $\psi \in [1,3]$
                \ENDFOR
                \STATE re-label $\{B(r_i|s_i,a_i) \leftarrow \hat{r}_\psi(s_i,a_i)\}^{|B|}_{i=1}$
            \ENDIF
            \ENDFOR
            \FOR{each gradient step} 
                \STATE sample minibatch $\{(s_i,a_i,s_{i+1},r_i)\} \sim B$
                \STATE optimize $J^{SAC}_{\theta}$, $J^{SAC}_{\phi}$, $J^{SAC}(\alpha)$
                \STATE $\hat{\theta}_i \leftarrow \tau\theta_i + (1-\tau)\hat{\theta}_i$ for $i \in [1,2]$
            \ENDFOR
        \ENDFOR
        \RETURN 
    \end{algorithmic}
\end{algorithm}

\begin{algorithm}[ht]
    \caption{\themis, Pre-training Phase}
    \label{alg:pre}
    \begin{algorithmic}[1]
        \STATE Initialize environment, buffer $B\leftarrow0$, temperature $\alpha$ and networks $\phi,\theta_1,\theta_2, \hat{\theta}_1, \hat{\theta}_2$
        \FOR{each episode}
            \STATE Reset environment and get $s_0$
            \FOR{each timestep $t$} 
                \STATE $a_t \sim \pi_\phi(a_t|s_t)$
                \STATE $s_{t+1} \leftarrow \mathcal{P}(s_t,a_t) $ \COMMENT{Sample transition from env}
                \STATE Calculate intrinsic reward $r^{int}_t \xleftarrow{} r^{int}(s_t)$
                \STATE $B \leftarrow B \cup \{(s_t,a_t,s_{t+1},r^{int}_t)\}$
            \ENDFOR
            \FOR{each gradient step} 
                \STATE sample minibatch $\{(s_i,a_i,s_{i+1},r^{int}_i)\} \sim B$
                \STATE Optimize $J^{SAC}_{\theta}$, $J^{SAC}_{\phi}$, $J^{SAC}(\alpha)$
                \STATE $\hat{\theta}_i \leftarrow \tau\theta_i + (1-\tau)\hat{\theta}_i$ for $i \in \{1,2\}$
            \ENDFOR
        \ENDFOR
        \STATE Evaluate $\pi_\phi$ and render environment
        \STATE Save $\pi_\phi$, $B$ in files
        \RETURN 
    \end{algorithmic}
\end{algorithm}

\section{Configuration options \& Hyper-parameters}
In table \ref{tab:coubia} we present the configurable parameters of $\themis$ along with the key options available to the user for conducting experiments. We should note that more parameters are available for the user to tweak depending on the level of detail and the parts of the system needed to configure. For parameters associated with the RL agent you can refer to table~\ref{tab:sacparam}.

\begin{table}[ht]
    \footnotesize
        \begin{tabular}{p{11em}p{16em}} 
        \toprule  
        \textbf{Parameter} & \textbf{Values}\\    
        \midrule
        Environment Domain & String\\
        Environment ID & String\\
        Action type & \{Discrete, Continuous\}\\
        State type & \{Tabular, Pixels, Grid \}\\
        Policy & \{MLP, CNN \}\\
        \midrule
        Unsupervised train steps & Positive Integer \\
        Training steps & Positive Integer \\
        Evaluation frequency & Positive Integer\\
        Evaluation episodes & Positive Integer\\
        Learn Reward & Boolean\\
        \midrule
        Ensemble size & Positive Integer\\ 
        Preference buffer capacity & Positive Integer\\
        Number of Interactions & Positive Integer\\
        Query batch size & Positive Integer\\
        Reward update frequency & Positive Integer\\
        Reward shaping & [scale: float, intercept: float]\\
        Sampling timespan & Positive Integer\\
        Sampling method & \{uniform, entropy, disagreement, k-center, k-center + entropy, k-center + disagreement\}\\
        Use human teachers & Boolean\\
        Synthetic Teacher specs & [5 parameters: floats]\\
        \midrule
        Explain Actions & Boolean \\
        Explain State Value  & Boolean \\
        Agent checkpoint frequency & Positive Integer\\
        Agent checkpoints location & Path string\\
        \bottomrule
        \end{tabular}
        \caption{Framework configuration parameters.}  
        \vspace{0.5cm}
        \label{tab:coubia}
\end{table}

In table \ref{tab:sacparam} we present the selected hyper-parameters for the SAC algorithm on our experiments while in table \ref{tab:hyperparam} there are the selected configuration parameters of $\themis$ for our experiments. For the most part, we selected the default values same or similar to the works that first introduced the components that is also used in our framework (eg. the CNN architecture of the RL agent, SAC's hyperparameters and reward model's hyperparameters).

\begin{table}[ht]
    \footnotesize
        \begin{tabular}{p{15em}p{12em}} 
        \toprule  
        \textbf{Parameter} & \textbf{Values}\\    
        \midrule
        Layers & 3 convolutional layers and 3 fully connected\\
        Convolutional channels per layer & [32, 64, 64]\\
        Convolutional kernels per layer & [8, 4, 1]\\
        Convolutional padding per layer & [0, 0, 0]\\
        Fully connected hidden layer units & [512, 512, action dim]\\
        Batch size & 512\\
        Replay buffer size & 500,000\\
        discount rate & 0.99 \\
        Learning iterations per round & 1\\
        Learning rate & 0.0005\\
        Optimizer & Adam\\
        Weight initializer & He\\
        Critic Target update frequency & 4 \\
        Loss & Mean squared error \\
        Initial random steps & 2,000 \\
        Entropy target  & $0.98*(-log(1/|A|))$\\
        \bottomrule
        \end{tabular}
        \caption{SAC hyper-parameters.}
        \vspace{0.5cm}
        \label{tab:sacparam}
\end{table}

\begin{table}[t]
    \footnotesize
        \begin{tabular}{p{16em}p{11em}} 
        \toprule  
        \textbf{Parameter} & \textbf{Values}\\    
        \midrule
        Environment Domain & ALE\\
        Environment ID & \{Breakout, Ms-Pacman, Pong\}\\
        Action type & Discrete\\
        State type & Pixels\\
        Policy & CNN\\
        \midrule
        Unsupervised train steps & 3,000 \\
        Training steps & 500,000 \\
        Evaluation frequency & 50 episodes\\
        Evaluation episodes & 10\\
        Learn Reward & True\\
        \midrule
        Ensemble size & 3\\ 
        Preference buffer capacity & 1,000\\
        Number of Interactions & 1400\\
        Query batch size & 128\\
        Reward update frequency & 200 episodes\\
        Reward shaping scale & float\\
        Reward shaping intercept & float\\
        Sampling time span & 50 episodes\\
        Sampling method & Max disagreement\\
        Use human teachers & False\\
        Synthetic teacher rationality & -1\\
        Synthetic teacher myopic Discount & 1\\
        Synthetic teacher mistake chance & 0\\
        Synthetic teacher skip chance & 0\\
        Synthetic teacher equal chance & 0\\
        \midrule
        Explain Actions & \{True, False\} \\
        Explain State Value  & \{False, True\} \\
        Agent checkpoint frequency & 500 steps\\
        \bottomrule
        \end{tabular}
        \caption{Configuration parameters for our experiments.}
        \vspace{0.5cm}
        \label{tab:hyperparam}
\end{table}

\section*{APPENDIX C: Exporting videos}
\begin{figure}[t]
    \centering
    \includegraphics[width=0.7\columnwidth]{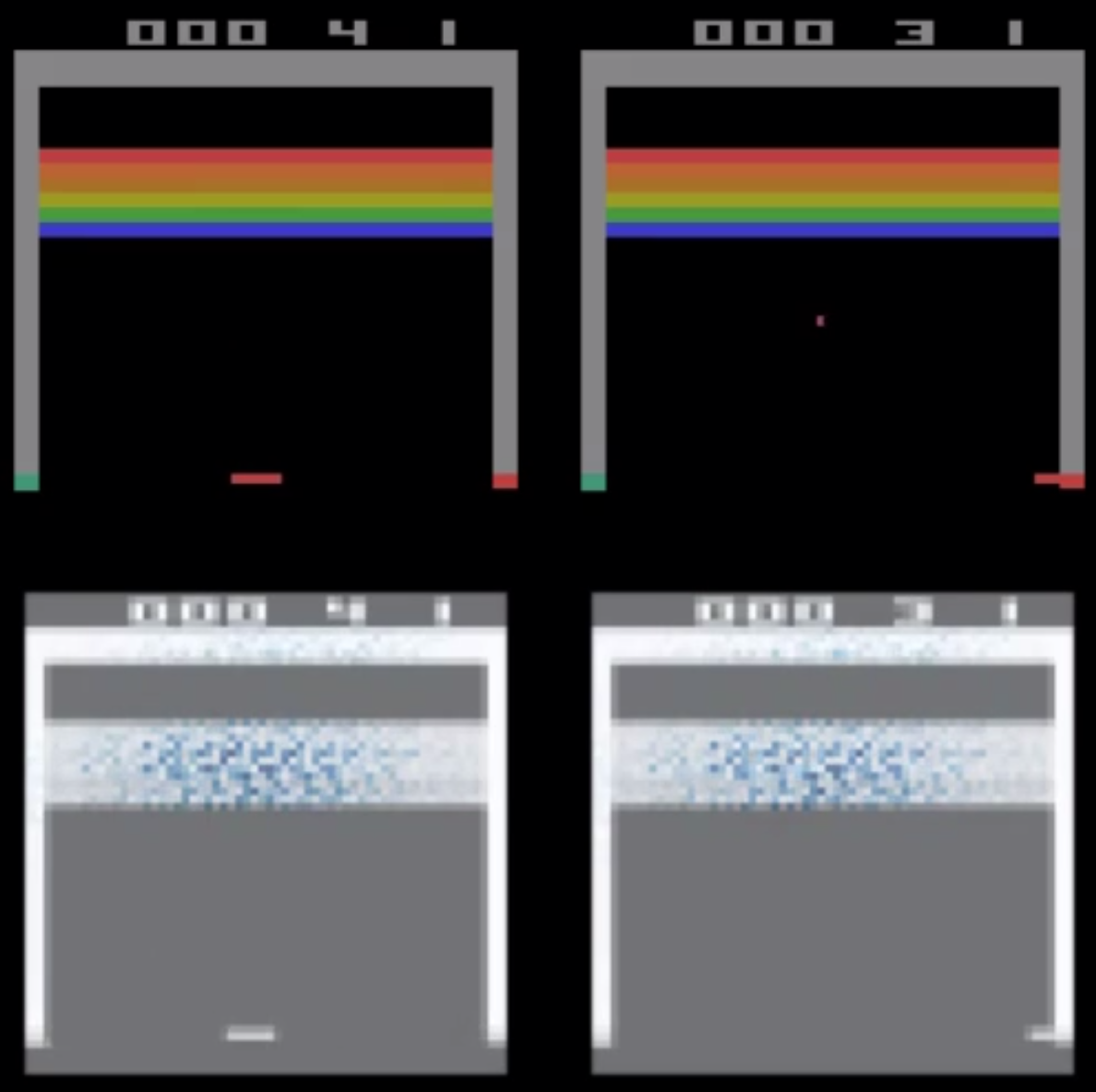}
    \caption{Snapshot of a clip export from $\themis$. The top row depicts the two segments rendered from the environment and the bottom has saliency map on their respective segments.\vspace{1.0cm}}
    \label{fig:clips}
\end{figure}

In the supplementary material, we include some video clips generated and exported using our framework. Here we explain what is portrayed in these clips and what each quarter represents. With figure \ref{fig:clips} as an example $\themis$ exports four clips. In our experiments we had the be exported as a single video clip with the four clips concatenated. The $y$ axis defines on which of the two pairs the clip refers to. On the left side is for segment 1 and on the right side for segment 2. Even without the clips on the bottom side these two clips should always be displayed. They are clips rendered from the two segments the users shall exert their preference on. They are sampled from the latest episodes of the replay buffer and specifically chosen based on the sampling rule selected for the experiment. For example with the maximum disagreement rule only clips with the biggest differences will be selected to display to the user. 

The bottom part of figure \ref{fig:clips} is associated with displaying explanatory information on the clips just above them. In the sample videos we provide we employ Integrated Gradients to produce saliency maps that are overlaid on the processed state that are provided as input to the agent's neural network model. Resizing the image is necessary to make the input to the CNNs in the agent use less data and that results in the lower resolution observed in the sample videos. The users can watch the clips as many times as they wish, and then they are tasked to make a selection between the left and right segment of the video. The whole interaction is facilitated through our web-based crowdsourcing platform. 

\end{document}